\documentclass[journal]{IEEEtran}

\usepackage{cite}
\usepackage[pdftex]{graphicx}
\usepackage{amsmath, amsthm}
\usepackage{newtxmath, newtxtext}
\usepackage{algorithm, algorithmic}
\usepackage{array}
\usepackage[caption=false,font=footnotesize]{subfig}
\usepackage{url}
\usepackage{hyperref}
\usepackage{xcolor}
\usepackage{enumitem}
\usepackage{color}
\usepackage{soul}
\usepackage{lineno}

\newlength\myindent
\newcommand\bindent{%
  \begingroup
  \setlength{\itemindent}{\myindent}
  \addtolength{\algorithmicindent}{\myindent}
}
\newcommand\eindent{\endgroup}

\newtheorem{proposition}{Proposition}

\graphicspath{{./images/}}
\DeclareGraphicsExtensions{.pdf,.jpeg,.png}

\begin{document}
\title{Neural Network Gaussian Process\\ Considering Input Uncertainty\\ for Composite Structures Assembly}

\author{Cheolhei Lee,~\IEEEmembership{}
        Jianguo Wu,~\IEEEmembership{Member,~IEEE},
        Wenjia Wang,~\IEEEmembership{}
        Xiaowei Yue,~\IEEEmembership{Member,~IEEE}
\thanks{C. Lee and X. Yue are with the Department of Industrial and
Systems Engineering, Virginia Tech, Blacksburg, VA, 24061.
(e-mail: cheolheil@vt.edu; xwy@vt.edu)}
\thanks{J. Wu is with the Department of Industrial Engineering and
Management, Peking University, Beijing.
(e-mail: j.wu@pku.edu.cn)}
\thanks{W. Wang is with Hong Kong University of Science and Technology, Clear Water Bay, Kowloon, Hong Kong.
(e-mail: wenjiawang@ust.hk)}
\thanks{
\textit{(Corresponding author: Xiaowei Yue)}}
\thanks{\copyright2020 IEEE. Personal use of this material is permitted.  Permission from IEEE must be obtained for all other uses, in any current or future media, including reprinting/republishing this material for advertising or promotional purposes, creating new collective works, for resale or redistribution to servers or lists, or reuse of any copyrighted component of this work in other works. }
}

\markboth{Accepted by IEEE/ASME TRANSACTIONS ON MECHATRONICS.}{}

\maketitle

\begin{abstract}
Developing machine learning enabled smart manufacturing is promising for composite structures assembly process. To improve production quality and efficiency of the assembly process, accurate predictive analysis on dimensional deviations and residual stress of the composite structures is required. The novel composite structures assembly involves two challenges: (i) the highly nonlinear and anisotropic properties of composite materials; and (ii) inevitable uncertainty in the assembly process. To overcome those problems, we propose a neural network Gaussian process model considering input uncertainty for composite structures assembly. Deep architecture of our model allows us to approximate a complex process better, and consideration of input uncertainty enables robust modeling with complete incorporation of the process uncertainty.  Based on simulation and case study, the NNGPIU can outperform other benchmark methods when the response function is nonsmooth and nonlinear. Although we use composite structure assembly as an example, the proposed methodology can be applicable to other engineering systems with intrinsic uncertainties. 
\end{abstract}


\IEEEpeerreviewmaketitle

\section{Introduction}
\IEEEPARstart{C}{omposite} materials are fabricated from two or more non-metallic, non-homogeneous constituent materials with different properties (such as carbon fiber and resin epoxy). Composite structures have increasingly replaced metals in many applications, such as aircraft, automobiles, civil structures, and sporting equipment in the last decades. For example, more than 50\% of a Boeing 787 is composed of composite materials for primary structures (e.g. fuselage, wings, tails, etc.) \cite{hale2006boeing}. The virtues of novel composite structures include high strength-to-weight ratio, high resistance to harsh chemicals, and better reliability.  Although composite structures have these advantages, high-quality and efficient assembly of them in the aircraft manufacturing process requires demanding endeavors. For example, due to multiple suppliers and manufacturing batches, composite structures are subject to inevitable dimensional deviations. Therefore, a pre-assembly dimensional shape control or shimming adjustment is usually conducted to reduce the dimensional gap between two composite structures \cite{wen2018feasibility, yue2018surrogate, wen2019virtual}. The current shape control is one of the most costly operations in aircraft assembly. Moreover, the quality of composite structures assembly in aerospace is directly linked to the safety of human life. Therefore, accurate predictive analysis for dimensional variations is essential to achieve the optimal manufacturing process and the ultra-high precision quality control of the product.

There are two important factors that should be considered in modeling dimensional shape control of composite structures: (i) composite structures have highly nonlinear and anisotropic properties; (ii) intrinsic uncertainty, which is associated with input, is involved in the process (e.g., from additive noise of actuator forces, heterogeneous parts, etc.). Therefore, a predictive model for composite structures should not only be highly expressive to accept the complex nature of composite structures, but also be capable of dealing with input uncertainty. Generally, finite element analysis (FEA) based models are considered as a classical method for the variation simulation \cite{fernlund2003finite, wen2018feasibility}. The advantage of FEA is that the model can provide reliable prediction based on physical mechanism with different levels of fidelity. However, FEA models have been restricted for the high computational cost, and the limitation becomes very prohibitive in the shape control of composite structures due to the real-time prediction requirement and the complex nature. Zhang \textit{et al.}\cite{zhang2016stream} developed a physics-based model for prediction of variation in composite parts assembly. Although the model considered noise from parts and fixture positions, it did not involve the dimensional shape control of composite structures, and it was lacking consideration of various uncertainties in the process.

Meanwhile, data-driven approaches have shown remarkable performance in manufacturing  fields due to development of machine learning and computation capabilities \cite{yuan2016bayesian, zhu2018tmech}. Among diverse machine learning methodologies, Gaussian processes (GPs) have been frequently used for surrogate models of FEA models in various fields \cite{taran2018}, due to their modeling efficiency, and capability of uncertainty quantification \cite{gpml2006}. Wan \textit{et al.}\cite{wan2017} used a GP regression model to predict deformation of assembly robots based on joint angles, so as to improve robot motion accuracy in a planned path of assembly process. However, their model cannot be used to the shape control modeling of composite structures assembly because the model did not consider the input noise. To cope with input noise in data for GPs, Cressie \textit{et al.}\cite{cressie2003} proposed the Kriging Adjusting for Location Error (KALE). They adjusted moments considering input location error, thereby producing a best linear unbiased predictor (BLUP) of the latent function. Yue \textit{et al.}\cite{yue2018surrogate} proposed the surrogate model considering uncertainties (SMU), which involves a GP and the other random terms associated with various uncertainty sources for the automated optimal shape control system for composite fuselage assembly. However, both the KALE and the SMU belong to shallow architecture models that may be insufficient to capture the complex nature of composite structures.

Deep architecture models are advantageous for approximating complex functions \cite{poole2016exponential}. As one of them, deep neural networks (DNNs) have shown remarkable performance in image recognition \cite{janssens2018}, and analysis of highly nonlinear problems \cite{kim2019}. Unfortunately, uncertainty quantification is still a challenging area in DNNs \cite{malinin2018predictive}, and most fully-connected feedforward DNNs presume that inputs are noise-free. Since Bayesian neural networks (BNNs) \cite{neal2012bayesian} provide predictive uncertainty by means of the Bayesian framework, BNNs can provide a meaningful predictive uncertainty when proper priors are provided. However, due to the intractable nature of the BNN's parameters, imposing such appropriate priors is challenging. Even though arbitrary priors are given (e.g., Gaussian priors), the resulted predictive uncertainty has nothing to do with input uncertainty. Furthermore, when the cost of sampling is very expensive such as in aircraft manufacturing, it is difficult to construct a reliable large network model with a limited data due to the high complexity of latent function. Therefore, naive DNNs and BNNs may be inappropriate for modeling composite structures assembly.

In recent years, there have been some efforts to connect GPs and DNNs to exploit their advantages \cite{damianou2013deep, neuralprocess2018}. Especially, upon the equivalence between GPs and infinite-width random DNNs \cite{neal1996, matthews2018gaussian}, some kernels for GPs to mimic infinite-width random DNNs have been proposed. Williams \cite{williams1997} introduced analytical kernels for single hidden-layer infinite-width NNs with Gaussian {function} and error function, and Cho and Saul\cite{cho2009} proposed an arc-cosine kernel, which is induced from heaviside step nonlinear functions of NNs, for a multiple-layer kernel machine. Lee \textit{et al.}\cite{lee2017} derived a GP that exactly corresponds to infinite DNNs using composite kernels, called neural network Gaussian process (NNGP), and showed that the NNGP outperforms finite-width full-connected networks in MNIST and CIFAR-10 data. Pang \textit{et al.}\cite{pang2019} compared the NNGP to shallow GPs for PDE regression problems and showed that the NNGP has higher expressivity than conventional shallow GP empirically.

There are several advantages in NNGP that DNNs do not have. First, they provide predictive uncertainty that is mostly meaningful in predictive modeling based analysis. Second, thanks to the derivation of NNGP from the infinite-width DNNs, it can exploit the high expressivity, while it does not require such a large data to get a satisfactory performance. However, current NNGPs assume that their inputs are noise-free like DNNs. It is a strong assumption for not only composite structures assembly, but also many of engineering applications such as machining, assembly, manufacturing, and  where input and output noise coexist. It is known that data-driven modeling without consideration of intrinsic input uncertainty may lead to a biased model and misinterpretation of the latent system \cite{carroll2006}. Therefore, to exploit those advantages in more realistic settings, considering input uncertainty in sophisticated systems is very important for achievement of ultra-high precision predictive modeling.

Motivated by limitations of the NNGP and other approaches for highly complex systems with input uncertainty, we propose a neural network Gaussian process considering input uncertainty (NNGPIU) in this paper with the following contributions.
\begin{itemize}[leftmargin=*]
    \item In order to address intrinsic input noise in complex systems, we propose a new way to consider input uncertainty with the NNGPIU, preserving the deep architecture of the NNGP, with a Monte-Carlo approximation scheme. The proposed NNGPIU method in our simulations and the case study of stress prediction in composite structures assembly outperforms the standard NNGP under the intrinsic input noise. The simulation study implies that our proposed approach also can be applied to other complex engineering systems subject to input noise. It has superior performance especially in the nonlinear system with nonsmooth functions and high-frequency functions. 
    \item We provide theoretical properties of our method. First, the NNGPIU is a best linear unbiased predictor of a function with input noise, and it asymptotically provides less predictive error than the standard NNGP. The computational algorithm is proposed and the computational cost is analyzed. The eigenspectrum analysis on the NNGPIU gives us some insights on how the NNGPIU behaviors differently to the shallow GPs.
\end{itemize}

The remainder of this paper is organized as follows. In Section II, we briefly review the NNGP, and derive the NNGPIU from a general model with input uncertainty. We also suggest an approximation scheme to realize the adjusted kernel of NNGPIU in detail, and discuss some theoretical analysis on the NNGPIU. Section III provides simulation studies of the NNGPIU and other benchmark methods. Two nonsmooth and high frequency functions are used to evaluate performance of our approach. Section IV applies the NNGPIU to modeling dimensional deviation and residual stress of composite fuselage for assembly to validate our approach with real data.

\section{Neural Network Gaussian Process Considering Input Uncertainty}
\subsection{Neural Network Gaussian Process (NNGP)}
In this section, we shortly review the NNGP without input uncertainty, and the equivalence between fully-connected random DNNs with multiple hidden-layers and GPs \cite{lee2017, pang2019}. Suppose that a DNN has $L$ hidden-layers, in which $L \ge 2$, and the number of units of $l$-th layer is $N_l$. The DNN accepts an input $x \in \mathbb{R}^d$, and returns an output $z^L \in \mathbb{R}$. Assume that the DNN has (i) independent and identically distributed (i.i.d.) weight and bias parameters of a network; and (ii) infinite number of units in layers (infinite-width). Let $\mathcal{X}$ be a convex and compact input space, and let $x,\,x'$ be independently sampled inputs from $\mathcal{X}$. We use $\phi(\cdot)$ to denote an activation function of the network, and $w_{ij}^l$ and $b_{i}^l$ to denote weight and bias parameters of $j$-th node in $l$-th hidden-layer, where $i$ stands for the index of node in the next layer. We assume that both $w_{ij}^l$ and $b^l_i$ are independent and identically Gaussian distributed with zero mean and finite variance, denoted by $\sigma_w^2/N_{l}$ and $\sigma_b^2$ respectively. $z_i^l(x)$ and $x_i^l(x)$ denote a post linear transformation and a post activation of $i$-th node in $l$-th layer associated with an original input $x$ respectively. Then, we can express a function of $l$-th hidden-layer (for $l=1,\,\ldots,\, L$) as follows.
\begin{align*}
    z_i^l(x) = b_i^{l} + \sum_{j=1}^{N_l}w_{ij}^l x_{j}^{l}(x)\,, \qquad x_j^l(x) = \phi(z_j^{l-1}(x)),
\end{align*}
where $x_j^l(x)$ are independent for $j\in{N_l}$, and $w_{ij}^l$ and $b_i^l$ are {independent normal}. Therefore, the resulting output $z_i^l(x)$ is the sum of {independent} random variables. Then, since $N_l$ is infinitely large, $z_i^l(x)$ becomes Gaussian by the central limit theorem (CLT). Note that we cannot apply the CLT to $z^0_i$'s, which is {the} post linear transformations of input layer, due to the finite width of input layer ($N_0 = d$). Instead, $z^0_i$ becomes Gaussian {because of} parameters with Gaussian prior, unlikely to other $z_i^l$ for {$l=1,\ldots,L$}. Let a finite samples from $\mathcal{X}$ be $\mathbf{X}=\{\mathbf{x}_1, \ldots, \mathbf{x}_T\}$. Consequently, $[z_i^l(\mathbf{x}_1),\ldots,z_i^l(\mathbf{x}_T]$ forms {an} $T$-dimensional multivariate Gaussian distribution, which is equivalent to a GP:
\begin{align*}
    z_i^l(\mathbf{X}) \sim \mathcal{GP}(\mathbf{0},\, c^l(\cdot,\, \cdot)),
\end{align*}
where $c^l$ is a corresponding covariance matrix. Note that the mean of GP is zero due to the zero mean of parameters. {The} covariance {of $z_i^l(x)$ and $z_i^l(x')$} ($c^l(x,\,x')$) is
\begin{align}
    c^l(x,\,x') &= \mathbb{E}\left[z_i^l(x)\,z_i^l(x')\right]\nonumber\\
            &= \sigma_b^2 + \sum_{j=1}^{N_l}\frac{\sigma_w^2}{N_l}\,\mathbb{E}\big[x_j^l(x)\,x_j^l(x')\big]\nonumber\\
            &= \sigma_b^2 + \sigma_w^2 \mathbb{E}\big[x_j^l(x)\,x_j^l(x')\big].\label{default_cov}
\end{align}
Let $\mathbb{E}\big[x_j^l(x)\,x_j^l(x')\big] = \mathbb{E}\big[x_j^l\,{x'}_j^{l}\big]$. Then, the expected value can be calculated as
\begin{align}
\mathbb{E}\big[x_j^l\,{x'}_j^{l}\big] &= \iint\phi(z_j^{l-1})\,\phi({z'}_j^{l-1})\,p(z_j^{l-1},\, {z'}_j^{l-1})\,dz_j^{l-1}{dz'}_j^{l-1}\nonumber\\
&\triangleq F_{\phi}\left(c^{l-1}(x, x),\, c^{l-1}(x', x'),\, c^{l-1}(x, x') \right).\label{complexkernel}
\end{align}
Setting a base covariance function for the input layer ($l=0$) as
\begin{align*}
    c^0(x,\,x') &= \sigma_b^2 + \frac{\sigma_w^2}{d} x\cdot x',
\end{align*}
$c^L$ can be obtained from recursive calculation of \eqref{default_cov} for $l=1,\cdots,L$ with a composite form. 

$F_\phi$ in \eqref{complexkernel} is determined by {the} activation function $\phi$. Lee \textit{et al.}\cite{lee2017} derived an analytic {form} of \eqref{default_cov} when $\phi$ is the rectified linear unit (ReLU:\, $\phi(x)=\max{(0,x)}$) based on Cho and Saul \cite{cho2009} as{,}
\setlength\arraycolsep{1.4pt}
\begin{eqnarray}
    c^l(x,\,x') &=& \sigma_b^2 + \frac{\sigma_w^2}{2\pi}\sqrt{c^{l-1}(x,x)c^{l-1}(x',x')}\nonumber\\
     &&\times \left(\sin{\theta^{l-1}_{x,x'}} + (\pi - \theta^{l-1}_{x,x'})\cos{\theta^{l-1}_{x,x'}}\right)\label{arccos},\\
    \theta^{l}_{x,x'} &=& \arccos{\left(\frac{c^l(x,x')}{\sqrt{c^l(x,x)\,c^l(x',x')}}\right)}.\nonumber
\end{eqnarray}
For the error activation function, Pang \textit{et al.}\cite{pang2019} derived {the} following analytic kernel.
\begin{gather}
    c^l(x,\,x') = \sigma_b^2 + \frac{2\sigma_w^2}{\pi}\,\theta^{l-1}_{x,x'},\label{arcsin}\\
    \theta^{l-1}_{x,x'} = \arcsin{\left(\frac{2c^{l-1}(x,x')}{\sqrt{(1+2c^{l-1}(x,x))(1+2c^{l-1}(x,x))}}\right)}.\nonumber
\end{gather}
In this paper, we refer to  \eqref{arccos} as arc-cosine kernel, and \eqref{arcsin} as arc-sine kernel.

Let $\mathbf{f}$ be observations on $\mathbf{X}$ so that we have training dataset $[\mathbf{X},\, \mathbf{f}].$ Then, the prediction for a test input $\mathbf{x}_\ast \notin \mathbf{X}$ of NNGP with $L$ hidden-layer can be obtained as 
\begin{align}
    \bar{f}_\ast &= \mathbf{C}^L_\ast\left(\mathbf{C}^L + \sigma_\epsilon^2 \mathbf{I}_n\right)^{-1}\mathbf{f},\label{nngp_mean}\\
    \mathbb{V}(f_\ast) &= c^L(x_\ast, x_\ast) - \mathbf{c}_\ast^L\left(\mathbf{C}^L + \sigma_\epsilon^2 \mathbf{I}_n\right)^{-1}\mathbf{c}^{L\top}_\ast, \label{nngp_var}
\end{align}
where $\mathbf{C}^L = c^L(\mathbf{X},\,\mathbf{X})$, $\mathbf{c}^L_\ast = c^L(x_\ast, \mathbf{X})$, $\sigma_\epsilon^2$ is the variance of observation noise ($\epsilon \sim \mathcal{N}(0,\,\sigma_\epsilon^2)$), and $\mathbf{I}_n$ is the $n \times n$ identity matrix.

\subsection{NNGPs with Input Uncertainty}\label{sec2.2}
\subsubsection{Statement of Composite Structures Assembly Problem}
{Dimensional deviations of composite structures involves intrinsic noise. Suppose that $p$ actuators are applied to adjust composite structures, and their forces constitute a $p$ dimensional vector $\mathbf{x}$ as input. The output associated with $\mathbf{x}$, denoted by $f(\mathbf{x})$, is the dimensional deviation of composite structures with binary directions that may be measured at a point of interest. Not surprisingly, the actuator force is subject to unobservable noise so that we observe the output associated with the noise-corrupted input. There are two main factors that brings the uncertainty of the actuators' forces} \cite{yue2018surrogate}: {(i) limited fabrication device tolerance of the actuators; and (ii) deviations of contact geometry of actuators. Suppose that these factors induce an additive noise, independent to input, and let the noise be $\mathbf{u}$ which is also a $p$ dimensional vector with a probability function $p_u(\cdot)$. While the input noise cannot be observed, characteristics of its distribution can be obtained from device instructions or inference from historical data as our prior knowledge.}

{Without loss of generality, dimensional deviations of composite structures with input uncertainty can be formulated as}
\begin{equation}
    y(\mathbf{x}) = f(\mathbf{x} + \mathbf{u}) + \epsilon,\quad \epsilon \sim \mathcal{N}(0,\, \sigma_\epsilon^2),\label{model}
\end{equation}
where $\epsilon$ is i.i.d. observation noise. Considering the complex nature of the latent function $f(\cdot)$, we assume that the target function $f$ is in the model space with the multi-layer covariance \eqref{model}. However, the standard NNGP ignores the effect of $u$, may induce large deviations in predictions, so an appropriate remedy must be devised in order to infer $f(\cdot)$ with given $\mathbf{x}$, $y(\mathbf{x})$, and our prior knowledge of the input noise.

\subsubsection{Adjusting Kernels}
To cope with the input uncertainty described in \eqref{model}, {we propose the NNGPIU that enables the standard NNGP to consider the input uncertainty. The input noise in the NNGP affects its kernel, which is the core of the NNGP. Accordingly, we adjust the kernel considering the effect of input noise by exploiting the prior information, and name the NNGP with the adjusted kernel as the NNGPIU. Assume that $f(\cdot)$ is the NNGP with $L$ layers and defined on a convex and compact space $\mathcal{X}$. That is,}
\begin{align*}
    f(x)\sim \mathcal{NNGP}(0,\, c^L(x, x')),\quad x,\,x'\in\mathcal{X},
\end{align*}
{in which the mean function is centralized to zero. Then, given a pair of observed inputs $x,\,x' \in \mathcal{X}$, we have}
\begin{align}
    \mathbb{C}\text{ov}[y(\mathbf{x}),\, y(\mathbf{x}')] &= \mathbb{C}\text{ov}[f(\mathbf{x} + \mathbf{u}),\,f(\mathbf{x}'+\mathbf{v})]\nonumber\\
		&= \mathbb{E}_{\mathbf{u,v}}[c^L(\mathbf{x + u,\, x'+v})]\nonumber\\
		&= \iint_{\mathbf{u,v}} c^L(\mathbf{x + u,\, x'+v})\,p_u(\mathbf{u})\,p_u(\mathbf{v})\nonumber\\
		&\triangleq k^L(x, x'),\label{nngpiu_k}
\end{align}
{where $k^L$ is defined as the adjusted kernel considering input uncertainty. For an unobserved input $x_\ast \in \mathcal{X}$, the adjusted kernel can be obtained from}
\begin{align*}
    k^L(\mathbf{x}_\ast,\,\mathbf{x}) &= \mathbb{C}\text{ov}[f(\mathbf{x}_\ast),\, y(\mathbf{x})]\\
    &= \mathbb{E}_{\mathbf{u}}[c^L(\mathbf{x}_\ast,\, \mathbf{x+u})]\\
    &= \int_\mathbf{u}c^L(\mathbf{x}_\ast,\,\mathbf{x+u}).
\end{align*}
Since $c^L$ is positive definite, $k^L$ will be also positive definite according to the Proposition 3.1. in the reference \cite{cervone2015} . {It turns out that the NNGPIU with the adjusted kernel is the BLUP of $f(\cdot)$ in terms of mean squared prediction error (MSPE).}
\begin{proposition}\label{prop1}
{The NNGPIU with $L$ hidden-layers, whose kernel is calculated by} \eqref{nngpiu_k}, {is a BLUP of $f(x_\ast)$ in terms of MSPE, where $\mathbf{x}_\ast$ is unobserved input.}
\begin{proof}
{Let a BLUP of $f(\mathbf{x}_\ast)$ be $\sum_{i=1}^n \hat{\beta}_i y(\mathbf{x}_i)\triangleq\hat{\boldsymbol{\beta}}^\top \mathbf{y}$ where $\hat{\boldsymbol{\beta}}=\arg\min_{\beta_i}\mathbb{E}\big[\big|f(x_\ast) - \sum_{i=1}^n \beta_i y(\mathbf{x}_i)\big|^2\big]$. Denote $\mathbb{E}[f(\mathbf{x}_\ast)^2]=c^L(\mathbf{x}_\ast,\,\mathbf{x}_\ast)$ as a constant $M$, then we have}
\begin{eqnarray}
    \mathbb{E}\big[\big\|f(x_\ast) - \boldsymbol{\beta^\top y}\big\|^2\big] &=& M + \boldsymbol{\beta}^\top \left(\mathbb{E}_{\mathbf{u,v}}\left[c^L(\mathbf{x}_i+\mathbf{u},\mathbf{x}_j+\mathbf{v})\right]_{ij}\right.\nonumber\\
    && \left.+\sigma_\epsilon^2\mathbf{I_n}\right)\boldsymbol{\beta} - 2\sum_{i=1}^n \beta_i \mathbb{E}_{u}\left[c^L(\mathbf{x}_\ast,\,\mathbf{x}_i+\mathbf{u})\right]\nonumber\\
    &=& M + \boldsymbol{\beta}^\top (\mathbf{K}^L+\sigma_\epsilon^2 \mathbf{I_n})\boldsymbol{\beta} - 2\boldsymbol{\beta}^\top \mathbf{k}^L_\ast,\nonumber
\end{eqnarray}
{where $\mathbf{K}^L$ and $\mathbf{k}^L_\ast$ are calculated by} \eqref{mckale}. {Then, its derivative with respect to $\boldsymbol{\beta}$ leads to $\hat{\boldsymbol{\beta}} = \mathbf{k}^L_\ast (\mathbf{K}^L+\sigma_\epsilon^2\mathbf{I_n})^{-1}$ of} \eqref{NNGPIUmean}.
\end{proof}
\end{proposition}

Unfortunately, there is no common distribution for the input noise allowing calculating $k^L$ analytically. Therefore, we need a numerical method to approximate \eqref{nngpiu_k}. In this paper, we use the Monte-Carlo (MC) approximation with the following scheme. 
% Suppose that $\mathbf{X}$ is a finite input samples from $\mathcal{X}$, and 
\begin{enumerate}
	\item Sample input {noises} $\{u_i\}_{i=1}^m$ from $p_u(\cdot)$ considering:
	\begin{itemize}
		\item the number of noise samples should be {sufficient} to make the maximum coefficient of variant to be less than 2.5\% \cite{cressie2003}.
		\item if data is standardized, then $p_u(\cdot)$ also must be adjusted consistently.
	\end{itemize}
	\item Use the sample to approximate kernels as follows.
	\begin{equation}
{\arraycolsep=1.5pt\def\arraystretch{1.8}
\left\{\begin{array}{l r}
k^l(\mathbf{x},\, \mathbf{x}') \approx \frac{1}{m^2}\sum_{i,j=1}^m c^l(\mathbf{x}+\mathbf{u}_i,\,\mathbf{x}'+\mathbf{u}_j), & \mathbf{x\ne x'},\\
k^l(\mathbf{x},\, \mathbf{x}) \approx \frac{1}{m}\sum_{i=1}^m c^l(\mathbf{x} + \mathbf{u}_i,\,\mathbf{x} + \mathbf{u}_i), & \\
k^l(\mathbf{x},\, \mathbf{x}_\ast) \approx \frac{1}{m}\sum_{i=1}^m c^l(\mathbf{x}+\mathbf{u}_i,\,\mathbf{x}_\ast), & \mathbf{x}_\ast \notin \mathbf{X}.
\end{array}\right.}
\label{mckale}
\end{equation}
\end{enumerate}
Note that \eqref{mckale} is not restricted to any specific distribution of input noise.

\subsubsection{Parameter Estimation}
{A convincing approach for estimation of parameters is maximum likelihood estimation. It is note worthy that assuming $f(\cdot)$ is a Gaussian process does not mean that $y(\mathbf{x})|\mathbf{x}$ is also Gaussian due to input uncertainty. Hence, the explicit estimation of parameters for the NNGPIU can be prohibitively expensive since the probability in the likelihood function requires another MC approximation. For this reason, we use maximum pseudo-likelihood estimation that requires no more than second moments like usual Gaussian likelihood. One of the attractive properties of pseudo-likelihood estimation is that it obtains the unbiased estimation of parameters. For more details about the properties, see} \cite{cressie2003, cervone2015}.

{Let $\theta$ be the parameters of $k^L$, let $\mathbf{K}^L$ be the gram matrix that consists of $k^L(\mathbf{x}_i, \mathbf{x}_j)$ for $i,\,j=1,\ldots,n$ and let $\mathbf{y}=[y(\mathbf{x}_1),\ldots,y(\mathbf{x}_n)]^\top$. The log pseudo-likelihood function of the NNGPIU is}
\begin{equation}
\ell(\theta) = -\frac{1}{2}\mathbf{y}^\top(\mathbf{K}^L+\sigma_\epsilon^2 \mathbf{I}_n)^{-1}\mathbf{y} - \frac{1}{2}\log{|\mathbf{K}^L+\sigma_\epsilon^2 \mathbf{I}_n|} + C,\label{lml}
\end{equation}
where $C$ is a redundant constant. Maximizing \eqref{lml} {with respect to $\theta$ with a gradient based method requires the derivative of $k^L$, which is analytically intractable. Thus, we need to calculate the derivative numerically. Additionally, maximizing $\ell(\theta)$ with a gradient method may suffer from local optima. Therefore, we need some exploration over the parameter space with different initialization.}

{It is also possible to estimate parameters with grid-search as} \cite{cho2009, lee2017}. {In some cases, it might be advantageous in computational time by skipping the numerical optimization, although this cross validation scheme may provide a heuristic estimation.}

\subsubsection{Prediction} 
{Once parameters are estimated, the predicted mean and variance of the NNGPIU for a new input $\mathbf{x}_\ast$ can be obtained by plugging $k^L$ into} \eqref{nngp_mean} and \eqref{nngp_var} as
\begin{align}
    \bar{f}_\ast &= {K}^L_\ast\,(\mathbf{K}^L + \sigma_\epsilon^2 \mathbf{I}_n)^{-1}\mathbf{y}\,\label{NNGPIUmean}\\
	\mathbb{V}(f_\ast) &= k^L(\mathbf{x}_\ast, \mathbf{x}_\ast) - \mathbf{k}^L_\ast\,(\mathbf{K}^L + \sigma_\epsilon^2 \mathbf{I}_n)^{-1}{\mathbf{k}^L_\ast}^\top,\nonumber
\end{align}
where $\mathbf{k}_\ast^L = k^L(\mathbf{x}_\ast, \mathbf{X})$, $\bar{f}_\ast$ is the mean, and $\mathbb{V}(f_\ast)$ is the variance. {Consequently, the best NNGPIU almost surely provides less MSPE than that of the NNGP as shown in Proposition} \ref{prop2}.
\begin{proposition}\label{prop2}
{There is an $\epsilon \ge 0$ such that}
\begin{align*}
    \normalfont \mathbb{E}[\|f(x_\ast) - \hat{f}_{NNGP}(x_\ast)\|_2^2 - \|f(x_\ast) - \hat{f}_{NNGPIU}(x_\ast)\|_2^2] \ge \epsilon,
\end{align*}
{where $\hat{f}_{NNGP}$ is the best NNGP model, and $\hat{f}_{NNGPIU}$ is the best NNGPIU. The equality holds if and only if $\mathbb{E}_{u,v}[c^L(x+u, x'+v)] = c^L(x, x')$, which means that the input noise has no effect on the kernel.}
\begin{proof}
{Suppose that we have the best NNGP and the best NNGPIU models as} \eqref{nngp_mean} and \eqref{NNGPIUmean} {respectively. For convenience, we omit the superscript $L$ in this proof. Then, the difference between MSPEs of them is}
\begin{eqnarray}
    \mathbb{E}[\|f(x_\ast)&-&c_\ast C^{-1}y\|_2^2 - \|f(x_\ast)-k_\ast K^{-1}y\|_2^2]\nonumber\\ &=& c_\ast C^{-1}KC^{-1}c_\ast - 2k_\ast C^{-1} + k_\ast K^{-1}k_\ast\label{mspe_diff}.
\end{eqnarray}
Let $T=C^{-1}KC^{-1}$ and $a = k_\ast C^{-1}$. Since $T$ and $T^{-1}$ are positive semidefinite (PSD), they have an unique square root PSD matrix respectively. Hence, \eqref{mspe_diff} becomes
\begin{align*}
    c_\ast Tc_\ast - 2ac_\ast + aT^{-1}a &= (c_\ast T^{1/2} - aT^{-1/2})^2 \ge 0.
\end{align*}
{The equality holds if $c_\ast T^{1/2} = aT^{-1/2}$, which is equivalent to $c_\ast C^{-1} = k_\ast K^{-1}$.}
\end{proof}
\end{proposition}
The overall procedure of NNGPIU is presented in Algorithm \ref{NNGPIU_algo}.

\begin{algorithm}
\caption{Neural Network Gaussian Process considering Input Uncertainty (NNGPIU)}
\label{NNGPIU_algo}
\begin{algorithmic}
    \STATE \textbf{Input}: $\mathbf{X}$, $\mathbf{y}$, $\mathbf{x}_\ast$
	\STATE \textbf{1. Initialization}\\
	\bindent
	\STATE Choose the number of hidden-layers in NNGPIU ($L$) \\
	\STATE Choose a maximum iteration for parameter estimation ($T$)\\
	\STATE Sample input noises from its distribution $p_u(\cdot)$\\
	\eindent
	\STATE \textbf{2. Parameter Estimation}\\
	\bindent
	\FOR{$k = [1:T]$}
	\STATE Set initial parameters $\theta_0^k$
	\STATE Set $\hat{\theta}^{i}=\arg{\max_\theta{\ell(\theta)}}$\\
	\bindent
	\STATE Calculate $K^L(\mathbf{X},\, \mathbf{X}\,|\,\theta)$ with $\{u_j\}_{j=1}^m$ based on \eqref{mckale}\\
	\eindent
	\STATE Save $\hat{\theta}^i$ and $\ell(\hat{\theta}^i)$
	\STATE $i = i + 1$
	\ENDFOR
	\STATE Set $\hat{\theta} = \hat{\theta}^i$ where $i=\arg{\max_i{\ell(\hat{\theta}^i)}}$
	\eindent
	\STATE \textbf{3. Prediction}\\
	\bindent
	\STATE Calculate $K^L(\mathbf{X},\,\mathbf{X}\,|\,\hat{\theta})$ and $k^L(x_\ast,\,\mathbf{X}\,|\,\hat{\theta})$\\
	\STATE $\bar{f}_\ast = K^L_\ast\left(K^L + \sigma_\epsilon^2 \mathbf{I}_n\right)^{-1}\mathbf{y}$\\
	\STATE $\mathbb{V}(f_\ast) = K^L(x_\ast, x_\ast) - K^L_\ast\left(K^L + \sigma_\epsilon^2\mathbf{I}_n\right)^{-1}K^{L\top}_\ast$
	\eindent
	\STATE \textbf{Return}: $\bar{f}_\ast$, $\mathbb{V}(f_\ast)$
\end{algorithmic}
\end{algorithm}

\subsection[title]{{Computational Cost}}
The deep architecture and the capability of considering input uncertainty of the NNGPIU come at an increased computational burden. The composite kernels multiply the computational cost with 3 to the power $L$, where $L$ is the number of layers, due to \eqref{complexkernel}. {Thus, if we pursue a deeper NNGPIU, then we must invest more time for training the model. The MC approximation scheme of} \eqref{mckale} {requires the square number of the input noise samples, so that the computational complexity for $n$ training samples and $m$ input noise samples require $\mathcal{O}(3^L m^2 k(\frac{1}{2}n^2))$. To alleviate the cost, we could exploit the formulation of composite kernel. For a pair of the same inputs, we only need to calculate} \eqref{complexkernel} {as }
\begin{align*}
    \mathbb{E}\big[x_j^l\,{x}_j^{l}\big] = F_{\phi}(c^{l-1}(x, x)),
\end{align*}
{since we don't need the degree between inputs. In this way, we can reduce the cost associated with diagonal entries of the gram matrix to $\mathcal{O}(3^L m^2 k(\frac{1}{2}n^2 - n))$.}

% \begin{table}[!t]
% 	\caption{Comparison of Computational Costs of Kernels ($n$: \# Samples)}
% 	\label{comp_cost}
% 	\centering
% 	\begin{tabular}{ccccc}
% 		ineine
% 		\textbf{Model} & \textbf{NNGP} & \textbf{KALE} & \textbf{NNGPIU}\\
% 		ine
% 		\textbf{Cost} & $\mathcal{O}(3^L k(\frac{1}{2}n^2))$ & $\mathcal{O}(k(\frac{1}{2}m^2 n^2))$ & $\mathcal{O}(3^Lk(\frac{1}{2}(n^2-n)))$\\
% 		ineine
% 	\end{tabular}
% \end{table}

\subsection[title]{{Eigenvalues of Composite Kernels}}
\begin{figure}[h]
	\centering
	\includegraphics[width=3.5in]{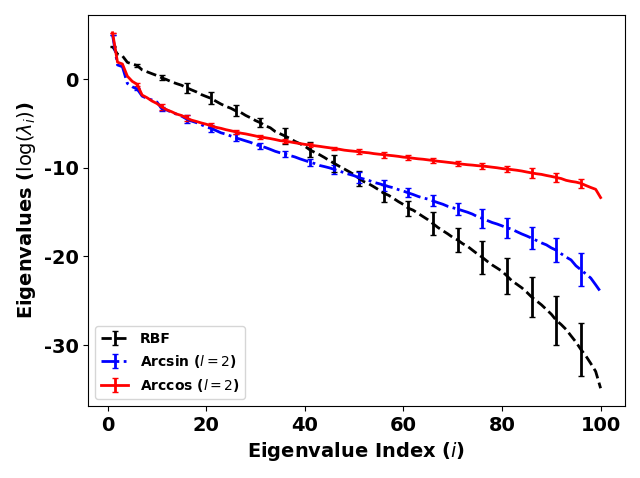}
	\caption{Eigenvalues of Kernels. 100 Gaussian samples are used, and the error bar is the result of 10 iterations. For the arcsine and arccosine kernels, $\sigma_b^2 = \sigma_w^2=1$ are used, and the length parameter of the RBF kernel is set with 1.}
	\label{spectrum}
\end{figure}

{Eigenvalues of a kernel model can be informative for getting insights into the model's behavior. For example, a kernel with slowly decaying eigenvalues is more suitable for rough functions} \cite{gpml2006}. {It is shown that eigenvalues of the Radial Basis Function (RBF) kernel with Gaussian inputs decay logarithmically so that eigenfunctions associated with large eigenvalues dominate the patterns of the model} \cite{zhu1998gaussian}. {In this paper, we empirically investigate eigenvalues of the NNGP (i.e., the arccosine and the arcsine kernels) with Gaussian inputs. Fig.} \ref{spectrum} {shows empirical eigenvalues of composite kernels and the RBF kernel in log scale.}

{Interestingly, decay rates of eigenvalues of the both composite kernels are still logarithmic, while the RBF kernel is not. This implies us that the NNGP and the NNGPIU's pattern tend to more rely on the leading eigenfunctions than GPs with the RBF kernel, yet preserving similar power for the rest eigenfunctions. It gives us an intuition that the NNGPIU may show more delicate balance between roughness and smoothness of the model. We observed that the dramatic logarithmic pattern of composite kernels is preserved even with different settings of parameters. Interestingly, the number of hidden-layers of the NNGPIU does not affect the decay rate of eigenvalues evidently.}

\section{Simulation Study}\label{sec3}
We implement the NNGPIU {for approximation of two functions} to observe its {performance.} A {zigzag} function and a near-square wave function are {approximated with the NNGPIU and benchmark methods: the shallow GPs with RBF and Mat\'ern kernels} \cite{gpml2006}, the NNGP \cite{lee2017}, and the KALE \cite{cressie2003}. {The shallow GPs are implemented with the scikit-learn package} \cite{scikit-learn}, {and the other methods are run with codes that we built in Python 3.8x. In the shallow GPs and the KALE, length and scale parameters are estimated, and the positive parameter of Mat\'ern kernel $\nu$ is fixed with 0.5 considering the patterns of the target functions. For the NNGP and the NNGPIU, weight and bias parameters are estimated, and two hidden-layers are used. Likelihood functions are maximized using the gradient method 'L-BFGS-B' in the scipy package} \cite{2020SciPy-NMeth}, {and the models are trained more than 10 times with random initial hyperparameters considering the local optima issue.}

{Training data is generated from both functions with additive input noise and observation noise. The input and observation noises are imposed with Gaussian distribution with zero mean and constant variance independently.} To emphasize {the} effect of input uncertainty, {the} variance of observation noise is set to be less than input {noise}'s. We iterated training and evaluation 20 times with different {samples} to reduce the sampling variability. For evaluation, the mean squared error (MSE) is calculated based on the true function over the entire input space. {Detailed settings are specified in each case.}

\subsection[title]{{Nonsmooth Function}}
The zigzag function has nondifferentiable points so we can observe how well the NNGPIU captures the pattern with input uncertainty. We set the input space to be an interval $[0,\,4]$, and 20 samples are generated from equispaced points with input noise. The variance of input noise is given with $\sigma_\mathbf{u}^2 = 0.1$, which is about 4\% of the interval length, and the observation noise is given with $\sigma_\epsilon^2 = 0.01$, which is 1\% of the maximum output value. 
\begin{figure}[h]
	\centering
	\includegraphics[width=3.3in]{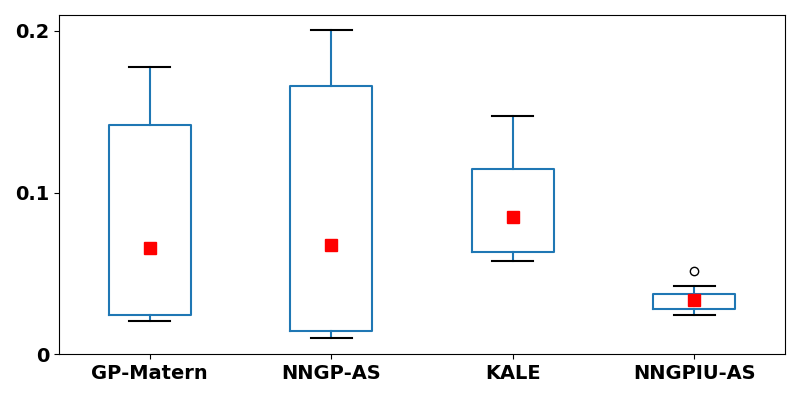}
	\caption{MSEs of 20 Iterations with the zigzag function. Red squares stand for averaged MSEs}
	\label{step_boxplot}
\end{figure}

\begin{table}[h]
	\caption{Averaged MSEs of 20 Iterations with the Zigzag Function}
	\label{step_table}
	\centering
	\begin{tabular}{ccccc}
		\hline\hline
		\textbf{Model} & \textbf{GP-Mat\'ern} & \textbf{NNGP} & \textbf{KALE} & \textbf{NNGPIU}\\
		\hline
		\textbf{MSE} & 0.0658 & 0.0678 & 0.0849 & 0.0338\\
		\hline\hline
	\end{tabular}
\end{table}

\begin{figure*}[h]
	\centering
	\subfloat[]{\includegraphics[width=3.5in]{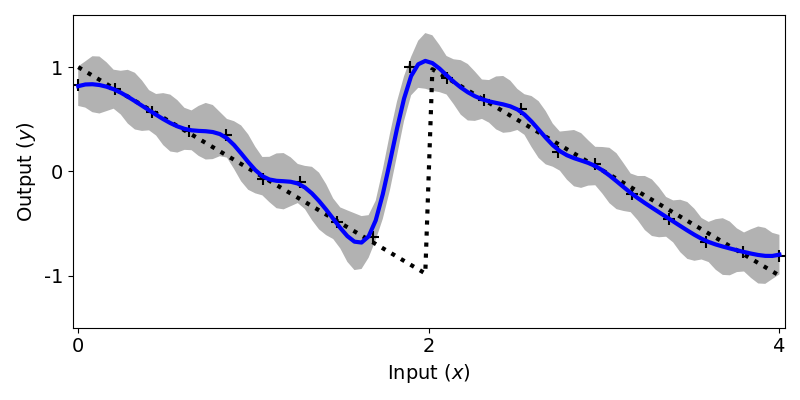}\label{step_matern}}\hfill
	\subfloat[]{\includegraphics[width=3.5in]{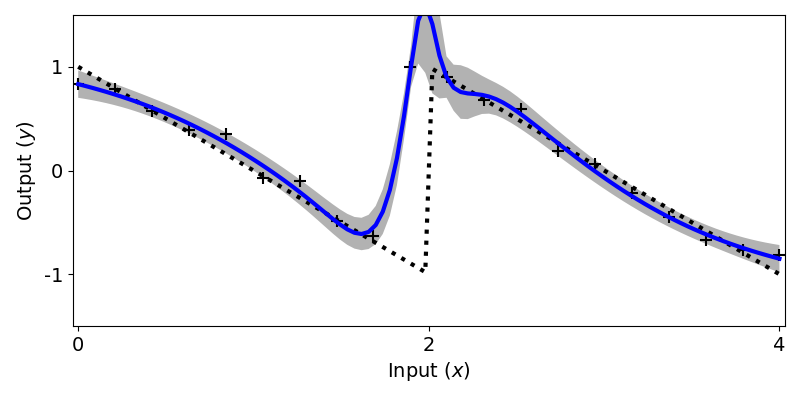}\label{step_nngp}}\\
	\subfloat[]{\includegraphics[width=3.5in]{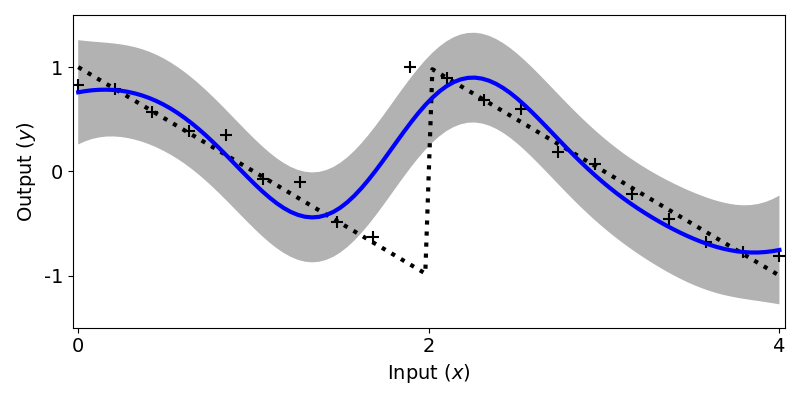}\label{step_kale}}\hfill
	\subfloat[]{\includegraphics[width=3.5in]{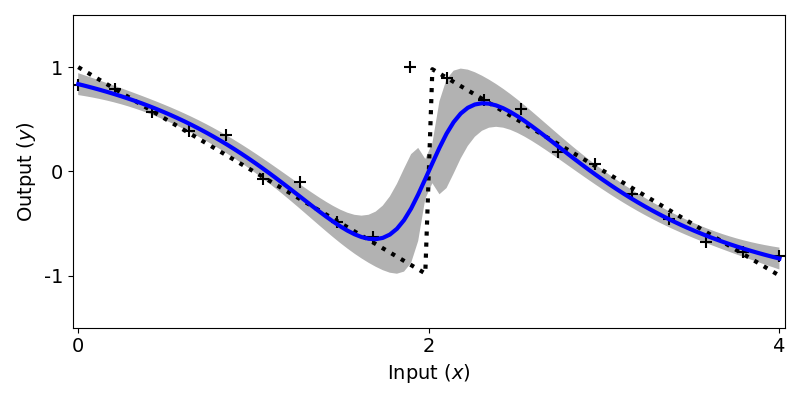}\label{step_NNGPIU}}
	\caption{Zigzag Function. (a) Shallow GP (Mat\'ern). (b) NNGP (Arcsine). (c) KALE. (d) NNGPIU (Arccosine). Dotted black lines are ground truth, and `+' symbols stand for training data. Each method is presented with the blue line as their mean and the shaded region as 95\% prediction interval.}
	\label{step}
\end{figure*}

Figure \ref{step_boxplot} is the boxplot of resulted MSEs of the models, and Table \ref{step_table} shows averaged MSEs of models. The shallow GP with the RBF kernel and the NNGPIU with the arccosine kernel are ignored since they were inferior to the arcsine kernel thereof. Apparently, the NNGPIU outperforms benchmark methods with the minimum prediction error and consistency. Fig. \ref{step} {shows one of the results of the methods.  The result demonstrates visually that the proposed NNGPIU works well for the zigzag function with a reliable prediction intervals. Note that both the shallow GP (Mat\'ern) and the NNGP are vulnerable to adversary data so that the models are overfitted. Interestingly, the KALE performed worse than the standard NNGP in terms of the averaged MSE over iterations.}

\subsection[title]{{High-Frequency Function}}
\begin{figure}[h]
	\centering
	\includegraphics[width=3.3in]{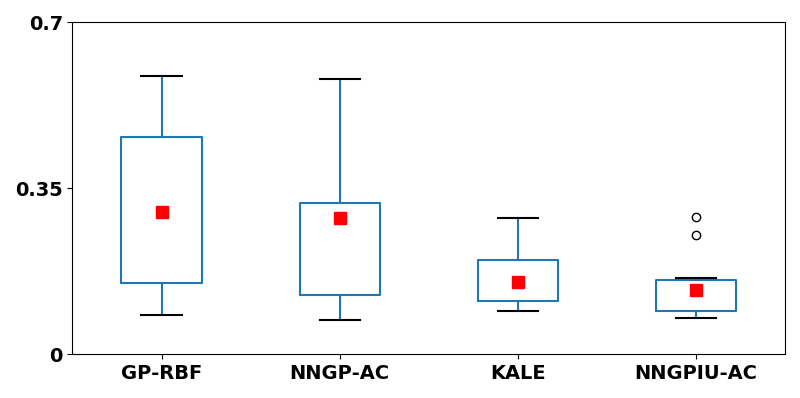}
	\caption{MSEs of 20 iterations with the near-square function.}
	\label{square_box}
\end{figure}
\begin{figure*}[h]
	\centering
	\subfloat[]{\includegraphics[width=3.5in]{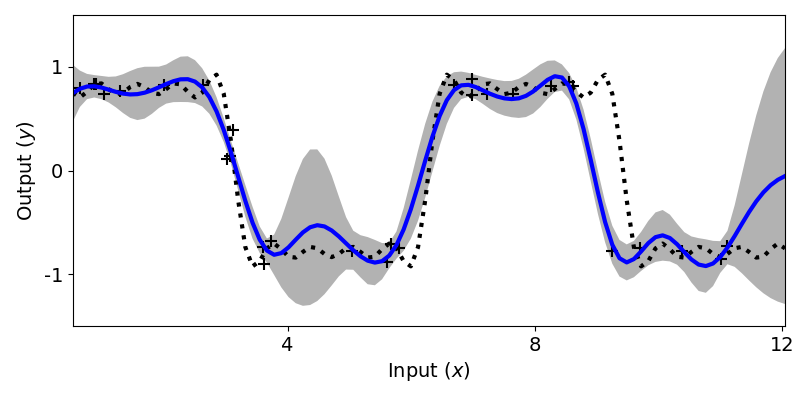}\label{square_gp}}\hfil
	\subfloat[]{\includegraphics[width=3.5in]{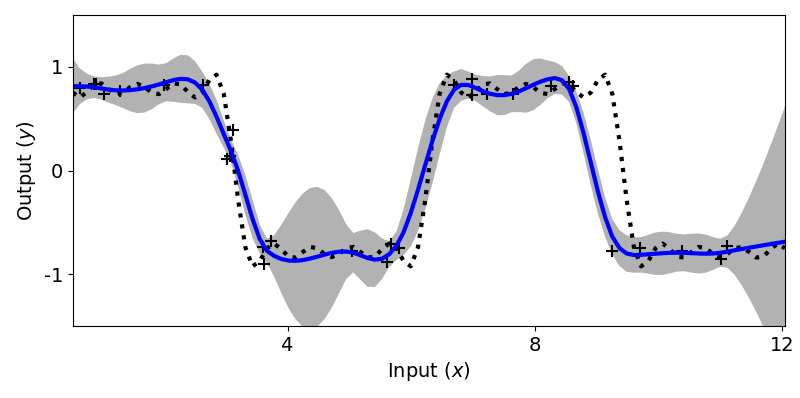}\label{square_matern}}\\
	\subfloat[]{\includegraphics[width=3.5in]{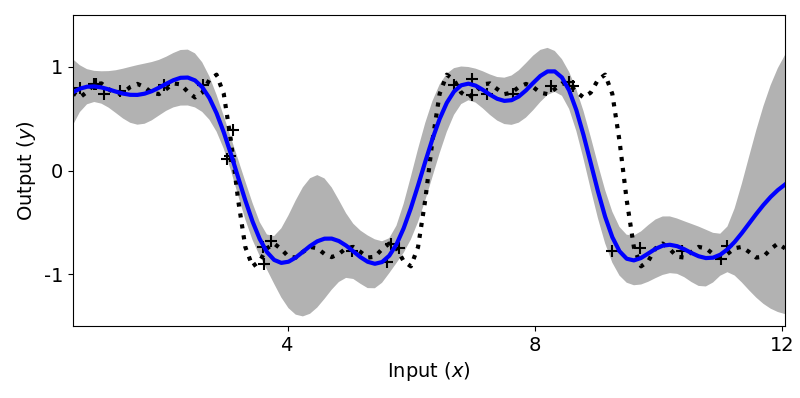}\label{square_kale}}\hfil
	\subfloat[]{\includegraphics[width=3.5in]{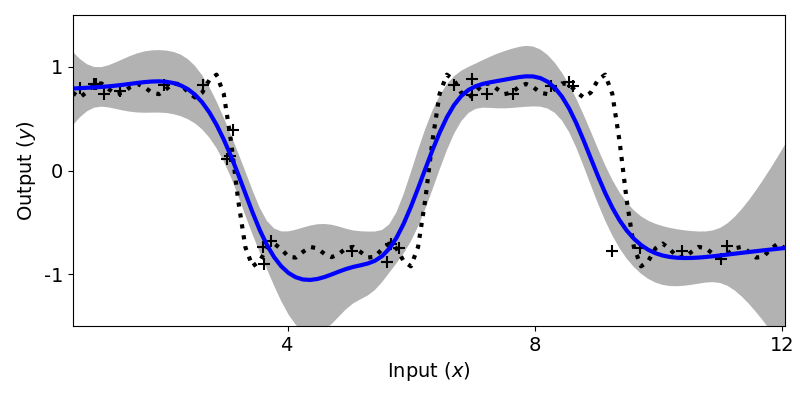}\label{square_NNGPIU}}
    \caption{Near-square wave function. (a) Shallow GP (RBF). (b) NNGP (Arccosine). (c) KALE. (d) NNGPIU (Arccosine).}
	\label{square}
\end{figure*}

{As a high frequency function, a near-square wave function is used. The near-square wave function is defined on an interval $\left[0, 4\pi\right]$, and 30 training samples are drawn uniformly over the interval. The input noise is imposed with variance $\sigma_u^2=0.03$, which is 2.5\%  of the interval-length, and the output noise is given with $\sigma_\epsilon=0.01$. Results are presented in Figure} \ref{square_box}, \ref{square} and Table \ref{square_table}. {The results of kernels with worse performance within a model are ignored.}

From the MSEs of compared models in Figure \ref{square_box} {and Table} \ref{square_table}, we can see that the NNGPIU outperforms the benchmark methods. However, in contrast to the nonsmooth case, it is difficult to recognize significant difference between compared models in Fig. \ref{square}. {The improvements in benchmark methods are induced by the smoothness of the target function and the reduction of $\sigma_u^2$. It is known that the RBF kernel is suitable for smooth functions as we have observed from its eigenvalues. Furthermore, the input noise is weakened so that performances of the shallow GP and the NNGP, which ignore input uncertainty, are improved.}

\begin{table}[h]
	\caption{Averaged MSEs of 20 Iterations with the Near-Square Wave Function}
	\label{square_table}
	\centering
	\begin{tabular}{ccccc}
		\hline\hline
		\textbf{Model} & \textbf{GP-RBF} & \textbf{NNGP-AC} & \textbf{KALE} & \textbf{NNGPIU-AC}\\
		\hline
		\textbf{MSE} & 0.2999 & 0.2873 & 0.1519 & 0.1355\\
		\hline\hline
	\end{tabular}
\end{table}

To summarize our simulation study, the NNGPIU has shown better performance than the benchmark methods for both target functions. This agrees with Proposition \ref{prop2} {and our conjecture from eigenvalues of composite kernels. The result of the simulation study shows that considering input uncertainty is critical for accurate function approximation with input noise. Especially, we observed that our approach is more advantageous than the KALE for nonsmooth and high frequency functions with consistent performance. Interestingly, the NNGP has shown a smaller averaged MSE than that of KALE in the zigzag function even without considering the input uncertainty. Thus, when the latent function is expected to has nonsmooth or high frequency, the NNGPIU can be a more promising choice.}

\section{Case Study: Composite Structures Assembly}\label{sec4}
We apply the NNGPIU for the composite structures assembly processes. In this case study, we consider two scenarios: (i) dimensional shape control of one single composite structure; (2) stress prediction in the composite structures assembly process. We use a well-calibrated finite element analysis (FEA) model, which has been validated by using the physical experiments. This FEA model can mimic real composite structures very accurately, from raw materials (carbon fiber and resin epoxy), ply design, and mechanical properties. More details about FEA model development refer to  \cite{yue2018surrogate,wen2018feasibility}. The FEA model has been calibrated via sensible variable identification and adjustment \cite{wang2019effective}.
\subsection{ Dimensional  Shape  Control  Of  Composite  Structure}
\begin{figure}[!t]
\centering
\subfloat[] {\includegraphics[width=1.6in]{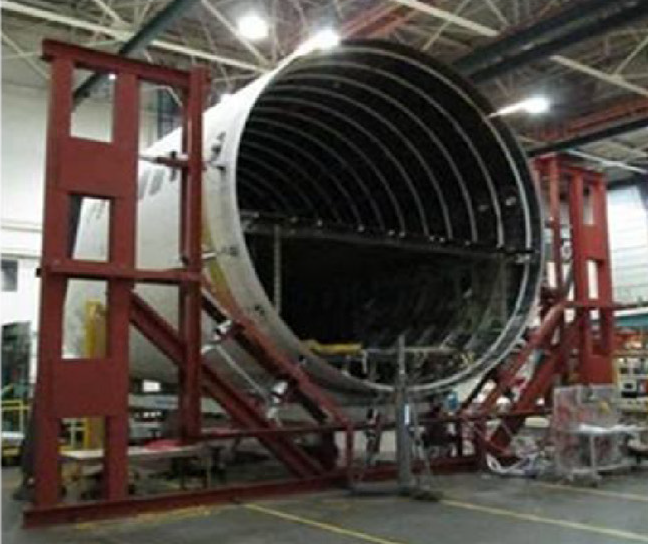}
\label{real}}
\subfloat[]{\includegraphics[width=1.4in]{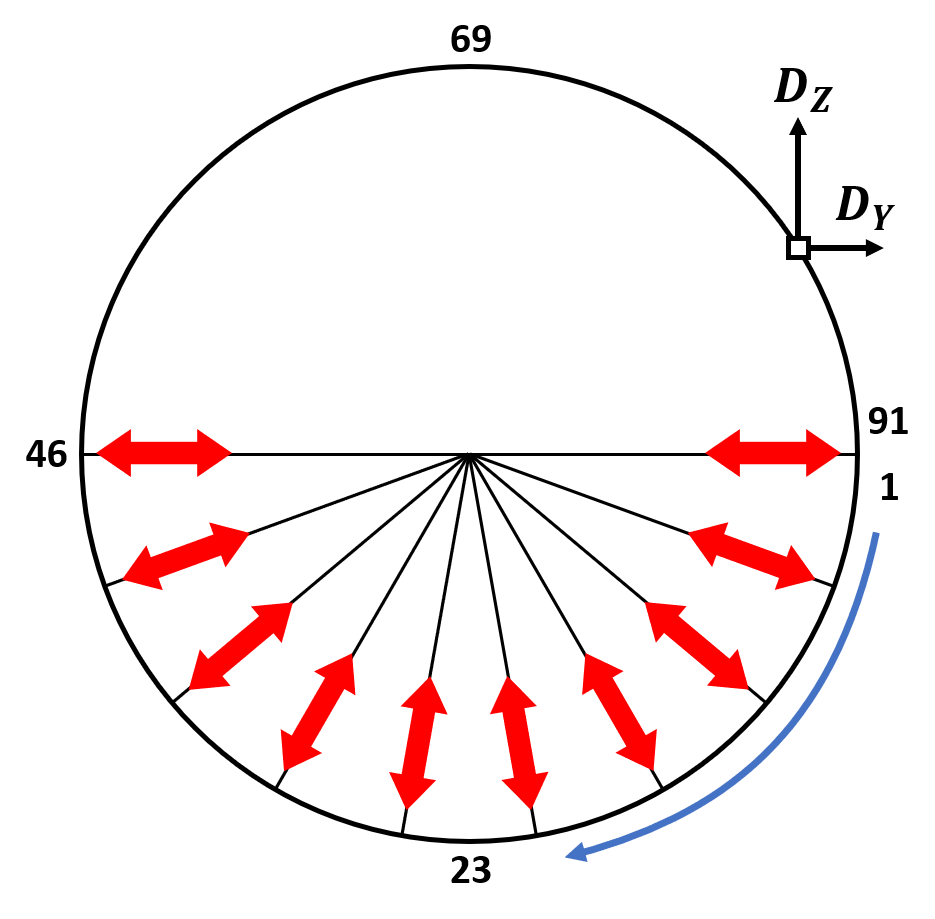}
\label{sketch}}
\caption{Composite structure assembly. (a) composite structure shape adjustment \cite{wen2018feasibility}. (b) Sketch for the adjustment of fuselage. Red arrows denote directions of 10 actuators' forces, and key points are numbered in counter-clockwise from 1 to 91. $D_Y$ and $D_Z$ show geometrical direction of deviation measurement.}
\label{fuselage}
\end{figure}

Dimensional shape control is a necessary step to reduce the gap between two structures before assembly. To  adjust dimensional deviations of one composite structure, Ten actuators are applied, as shown in Fig. \ref{real}. Actuators' forces are set with a range of $\pm 600$ lbf. Dimensional deviations are measured with two orthogonal directions (referred as $D_Y$ and $D_Z$) in inch scale at 91 key points that are linearly positioned around the composite structure as shown in Fig. \ref{sketch}. A total deformation at $i$-th key point ($D^{(i)}$) is recalculated as
\begin{equation}
    D^{(i)} = \sqrt{{D_Y^{(i)}}^2 + {D_Z^{(i)}}^2},\quad i=1,\ldots,91.
    \label{totaldef}
\end{equation}

We build models for each direction and key point, using a multivariate regression term to capture the linear pattern and a stochastic process term to learn the nonlinear pattern as follows.
\begin{align}
{\arraycolsep=1.5pt\def\arraystretch{1.8}
\left\{
    \begin{array}{c}
    \mathbf{D}_Y^{(i)}(\mathbf{X})= \mathbf{X\beta}_Y + \mathbf{Z}_Y^{(i)}(\mathbf{X})\\
    \mathbf{D}_Z^{(i)}(\mathbf{X})= \mathbf{X\beta}_Z + \mathbf{Z}_Z^{(i)}(\mathbf{X})
    \end{array}\right.},\quad i=1,\ldots,91.
    \label{casemodel}
\end{align}
where $\mathbf{X} \in \mathbb{R}^{10}$ is a row vector of actuators' forces, $\mathbf{\beta} \in \mathbb{R}^{10}$ is a vector of linear predictor, and $\mathbf{Z}^{(i)}$ is a random process. The multivariate regression term $\mathbf{X\beta}$ is used to capture the linear relationship between actuators' forces and dimensional deviations, and the stochastic term $\mathbf{Z}^{(i)}(\mathbf{X})$ will be trained for nonlinear patterns of composite structures' deformation. In this scenario, we use the proposed NNGPIU method as well as several benchmark methods including a shallow GP with the RBF kernel, NNGP, KALE  for the stochastic term.

Fifty shape control samples are generated for training and thirty samples are generated for testing respectively by using the Maximin Latin hypercube design \cite{dace2018}. Sequential experimental design used the active learning for Gaussian process considering uncertainties \cite{yue2020active}. The actuators' noise follows the Gaussian distribution with zero mean and constant variance. The variance is set with 0.5\% of the maximum force of actuators (about 3.0 lbf) based on the actuator manual and engineering practice. Without losing generality, we assume other uncertainties can be represented with the input uncertainty level we set.
Models are trained and tested with training and testing dataset respectively. The same hyperparameters are used for each model, and models are trained ten times with differently initialized hyperparameters. For the NNGP and NNGPIU, we use the arc-sine kernel, since we have observed that the arc-sine kernel perform better than the arc-cosine kernel in this case study.  Each model is evaluated with mean absolute prediction error (MAE).

% \begin{figure*}[!t]
%	\centering
%	\subfloat[]{\includegraphics[width=3in]{gp_test.png}\label{kale}}\hfil
%	\subfloat[]{\includegraphics[width=3in]{nngp_test.png}\label{fea_nngp_test}}\\
%	\subfloat[]{\includegraphics[width=3in]{kale_test.png}\label{fea_kale_test}}\hfil
%	\subfloat[]{\includegraphics[width=3in]{nngpiu_Test.png}\label{fea_nngpiu_test}}
%	\caption{Mean absolute prediction error of five models for composite fuselage assembly. (a) Shallow GP (Mat\'ern). (b) NNGP. (c) KALE. (d) NNGPIU.}
%	\label{fea_test}
% \end{figure*}

\begin{table}[!t]
\renewcommand{\arraystretch}{1.3}
	\caption{Mean Absolute Prediction Error of Models for Composite Structure Shape Control}
	\label{mae_fea}
	\centering
	\begin{tabular}{cccccc}
		\hline\hline    
		\textbf{Model} & \textbf{Linear} & \textbf{Shallow GP} & \textbf{NNGP} & \textbf{KALE} & \textbf{NNGPIU}\\
		\hline
		\textbf{MAE} & 0.00937 & 0.00942 & 0.01084 & 0.00929 & 0.00935\\
		\hline\hline
	\end{tabular}
\end{table}

The MAEs of all the methods are summarized in Table \ref{mae_fea}. According to the Table \ref{mae_fea}, all the methods have quite comparable performance in prediction of dimensional shape control. One main reason is that when the input uncertainty is very small, the linear pattern dominates in the dimensional shape control. The shallow GP and NNGP have marginally greater prediction errors than the others. NNGPIU and KALE provide marginally smaller prediction errors by considering input uncertainty. According to the conclusions from simulation study, We have seen that the KALE has shown comparable performance to the NNGPIU's when it is the near-square wave function, while NNGPIU performs much better than KALE and other benchmark methods when it is the zigzag function.  We conjecture that the performance of NNGPIU becomes more significantly advantageous when the latent process becomes more non-smooth.  Based on this scenario, we conclude that both KALE and NNGPIU can provide very accurate prediction for composite structures shape control with input uncertainty, where linear pattern dominates the hidden data pattern. 

\begin{figure}[!t]
	\centering
	\includegraphics[width=3in]{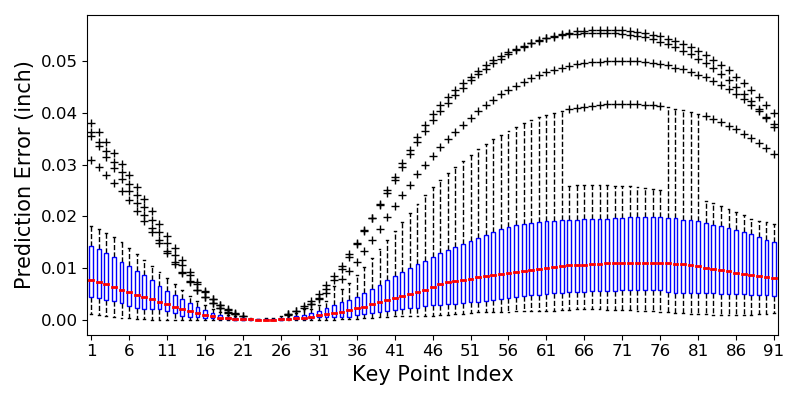}
	\caption{Prediction error of NNGPIU}
	\label{nngpiu_test}
\end{figure}
Fig. \ref{nngpiu_test} is a boxplot of NNGPIU' residuals for the test dataset. The $X$-axis stands for the index of key points , and $Y$-axis stands for a total deformation in key points, that is calculated {by} \eqref{totaldef}. Since actuators are installed under the lower part of the composite structure (Fig. \ref{sketch}), we can see that the lower part tends to have less prediction error than the upper part, which is consistent with the physical experiments.

\subsection{Stress Prediction in Composite Structures Assembly}
After composite structures assembly and rivet joins, residual stresses may remain after the release of fixtures and actuators. The residual stress can result in severe quality and reliability issue in the assembled product. Therefore, it is critical to do residual stress prediction and analysis. In this subsection, we investigate the virtual assembly simulation of composite structures and test the performance of the developed model with incorporating input uncertainties. 

\begin{figure}[!t]
	\centering
	\includegraphics[width=3in]{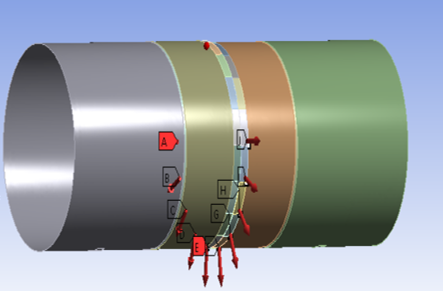}
	\caption{Virtual assembly simulation of composite structures}
	\label{assembly_FEA}
\end{figure}

\begin{figure}[!t]
	\centering
	\includegraphics[width=3in]{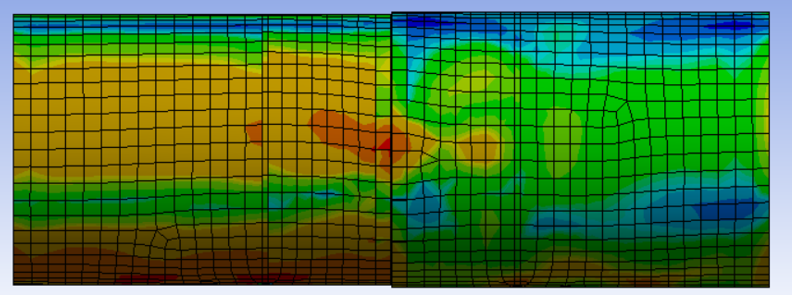}
	\caption{Residual stress after releasing in Virtual assembly simulation }
	\label{assembly_stress}
\end{figure}

\begin{figure}[!t]
	\centering
	\includegraphics[width=3.3in]{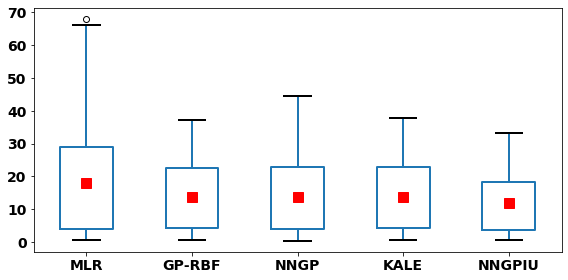}
	\caption{MAE Boxplots of Models for Residual Stress Prediction during Composite Structures Assembly}
	\label{assembly_boxplot}
\end{figure}

\begin{table}[!t]
\renewcommand{\arraystretch}{1.3}
	\caption{Mean Absolute Prediction Error of Models for Residual Stress Prediction during Composite Structures Assembly}
	\label{mae_fea2}
	\centering
	\begin{tabular}{cccccc}
		\hline\hline    
		\textbf{Model} & \textbf{Linear} & \textbf{Shallow GP} & \textbf{NNGP} & \textbf{KALE} & \textbf{NNGPIU}\\
		\hline
		\textbf{MAE} & 18.143 & 13.668 & 13.619 & 13.742 & 11.884\\
		\hline\hline
	\end{tabular}
\end{table}

Fig. \ref{assembly_FEA} is a schematic diagram of the composite structures assembly, where ten actuators' forces are used to adjust the dimensional deviations of each structure. The virtual assembly includes five steps: (i) generate the design-shaped composite structures; (ii) introduce the manufacturing deviations to the design-shaped structures and generate manufactured structures; (iii) conduct the dimensional shape control to eliminate the dimensional gap between two composite structures; (iv) rivet joins to bond two composite structures; (v) release the actuators and spring-back of two composite structures. More details related to virtual assembly refer to \cite{wen2019virtual}. After virtual assembly, residual stress can be generated. One example of residual stress map of two composite structures can be found in  Fig. \ref{assembly_stress}. 

We generated thirty virtual assembly samples for training and twenty virtual assembly samples for testing. For every virtual assembly, we have ten actuators' forces for each composite structure. The output includes residual stress from 128 critical points. The MAEs of stress prediction via multiple methods are summarized in Table \ref{mae_fea2} and Fig. \ref{assembly_boxplot}. We can find the multivariate linear regression can realize prediction error 18.143 psi, while the shallow GP with RBF kernel, NNGP, and KALE can realize 13.668 psi, 13.619 psi and 13.742 psi, respectively. That means nonlinear patterns exists in the composite structures assembly. Only via linear terms, the model cannot realize accurate prediction. The proposed NNGPIU has the best residual stress prediction, with an MAE 11.884 psi. It indicates the NNGPIU can outperform other benchmark methods by incorporating input uncertainties. The NNGPIU is more advantageous than the benchmark methods such as GP, NNGP and KALE, when the latent pattern is nonlinear and nonsmooth.  

We summarize the two scenarios in case study. (1) In dimensional shape control of composite structure, it is a single-shot force-deformation process, so the linear patterns dominates. The NNGPIU and other benchmarks can realize comparable performance, and NNGPIU and KALE are marginally better. (2) In virtual assembly of two composite structures, the structures need to go through multiple steps, including shape control, rivet joins, release and spring-back. The response surface is nonsmooth and nonlinear. The NNGPIU can realize the best prediction performance than other linear and nonlinear benchmark methods. The case study result is consistent with the simulation study.

\section{Conclusion}\label{sec5}
In this paper, motivated by unique characteristics of composite structures and inevitable uncertainty in assembly process, we suggested necessity of a deep architecture model considering input uncertainty. To exploit advantages of GPs and DNNs, we have derived a neural network Gaussian process considering input uncertainty (NNGPIU) to achieve more accurate prediction of composite structures assembly. In the derivation of our method, we have shown that the NNGPIU is the bear linear unbiased predictor of a target function with input uncertainty, and the best NNGPIU's MSPE is less than the best NNGP model's. Furthermore, the smoothness of the NNGPIU model is investigated with eigenvalues of composite kernels.

After deriving our method, we implemented the NNGPIU and other benchmark methods on two simulation data, and observed that {the NNGPIU can achieve better performance than the NNGP and the KALE for the nonsmooth and high frequency functions.} In the case study, we considered two scenarios: dimensional shape control of one composite structure, and residual stress prediction for virtual assembly of two composite structures assembly. In the shape control of composite structure, the NNGPIU and the KALE can realize comparable performance due to the linear pattern dominates the response function. While in virtual assembly of composite structures, the NNGPIU can realize the optimal residual stress prediction than other linear and nonlinear benchmark methods.  Based on the simulation study and case study, we conclude that dimensional deviations of composite structure have smooth response surface with respect to to actuators' forces, while the residual stress in virtual composite structures assembly has nonsmooth and nonlinear response surface. Consequently, the NNGPIU can be more advantageous than the standard NNGP and the KALE for nonsmooth and nonlinear functions with input uncertainty.

\bibliographystyle{IEEEtran}
\bibliography{nngp_trans}

\begin{thebibliography}{10}
\providecommand{\url}[1]{#1}
\csname url@samestyle\endcsname
\providecommand{\newblock}{\relax}
\providecommand{\bibinfo}[2]{#2}
\providecommand{\BIBentrySTDinterwordspacing}{\spaceskip=0pt\relax}
\providecommand{\BIBentryALTinterwordstretchfactor}{4}
\providecommand{\BIBentryALTinterwordspacing}{\spaceskip=\fontdimen2\font plus
\BIBentryALTinterwordstretchfactor\fontdimen3\font minus
  \fontdimen4\font\relax}
\providecommand{\BIBforeignlanguage}[2]{{%
\expandafter\ifx\csname l@#1\endcsname\relax
\typeout{** WARNING: IEEEtran.bst: No hyphenation pattern has been}%
\typeout{** loaded for the language `#1'. Using the pattern for}%
\typeout{** the default language instead.}%
\else
\language=\csname l@#1\endcsname
\fi
#2}}
\providecommand{\BIBdecl}{\relax}
\BIBdecl

\bibitem{hale2006boeing}
J.~Hale, ``Boeing 787 from the ground up,'' \emph{Aero}, vol.~4, no.~24, p.~7,
  2006.

\bibitem{wen2018feasibility}
Y.~Wen, X.~Yue, J.~H. Hunt, and J.~Shi, ``Feasibility analysis of composite
  fuselage shape control via finite element analysis,'' \emph{Journal of
  Manufacturing Systems}, vol.~46, pp. 272--281, 2018.

\bibitem{yue2018surrogate}
X.~Yue, Y.~Wen, J.~H. Hunt, and J.~Shi, ``Surrogate model-based control
  considering uncertainties for composite fuselage assembly,'' \emph{Journal of
  Manufacturing Science and Engineering}, vol. 140, no.~4, p. 041017, 2018.

\bibitem{wen2019virtual}
Y.~Wen, X.~Yue, J.~H. Hunt, and J.~Shi, ``Virtual assembly and residual stress
  analysis for composite fuselage assembly process,'' \emph{Journal of
  Manufacturing Systems}, vol.~52, pp. 55--62, 2019.

\bibitem{fernlund2003finite}
G.~Fernlund, A.~Osooly, A.~Poursartip, R.~Vaziri, R.~Courdji, K.~Nelson,
  P.~George, L.~Hendrickson, and J.~Griffith, ``Finite element based prediction
  of process-induced deformation of autoclaved composite structures using 2d
  process analysis and 3d structural analysis,'' \emph{Composite Structures},
  vol.~62, no.~2, pp. 223--234, 2003.

\bibitem{zhang2016stream}
T.~Zhang and J.~Shi, ``Stream of variation modeling and analysis for compliant
  composite part assembly—part i: single-station processes,'' \emph{Journal
  of Manufacturing Science and Engineering}, vol. 138, no.~12, p. 121003, 2016.

\bibitem{yuan2016bayesian}
Y.~Yuan, H.-T. Zhang, Y.~Wu, T.~Zhu, and H.~Ding, ``Bayesian learning-based
  model-predictive vibration control for thin-walled workpiece machining
  processes,'' \emph{IEEE/ASME transactions on mechatronics}, vol.~22, no.~1,
  pp. 509--520, 2016.

\bibitem{zhu2018tmech}
K.~{Zhu} and Y.~{Zhang}, ``A cyber-physical production system framework of
  smart cnc machining monitoring system,'' \emph{IEEE/ASME Transactions on
  Mechatronics}, vol.~23, no.~6, pp. 2579--2586, 2018.

\bibitem{taran2018}
N.~{Taran}, D.~M. {Ionel}, and D.~G. {Dorrell}, ``Two-level surrogate-assisted
  differential evolution multi-objective optimization of electric machines
  using 3-d fea,'' \emph{IEEE Transactions on Magnetics}, vol.~54, no.~11, pp.
  1--5, Nov 2018.

\bibitem{gpml2006}
C.~K. Williams and C.~E. Rasmussen, \emph{Gaussian processes for machine
  learning}.\hskip 1em plus 0.5em minus 0.4em\relax MIT press Cambridge, MA,
  2006.

\bibitem{wan2017}
A.~{Wan}, J.~{Xu}, H.~{Chen}, S.~{Zhang}, and K.~{Chen}, ``Optimal path
  planning and control of assembly robots for hard-measuring easy-deformation
  assemblies,'' \emph{IEEE/ASME Transactions on Mechatronics}, vol.~22, no.~4,
  pp. 1600--1609, Aug 2017.

\bibitem{cressie2003}
N.~Cressie and J.~Kornak, ``Spatial statistics in the presence of location
  error with an application to remote sensing of the environment,''
  \emph{Statistical Science}, pp. 436--456, 2003.

\bibitem{poole2016exponential}
B.~Poole, S.~Lahiri, M.~Raghu, J.~Sohl-Dickstein, and S.~Ganguli, ``Exponential
  expressivity in deep neural networks through transient chaos,'' in
  \emph{Advances in Neural Information Processing Systems}, 2016, pp.
  3360--3368.

\bibitem{janssens2018}
O.~{Janssens}, R.~{Van de Walle}, M.~{Loccufier}, and S.~{Van Hoecke}, ``Deep
  learning for infrared thermal image based machine health monitoring,''
  \emph{IEEE/ASME Transactions on Mechatronics}, vol.~23, no.~1, pp. 151--159,
  Feb 2018.

\bibitem{kim2019}
D.~{Kim}, J.~{Kwon}, S.~{Han}, Y.~{Park}, and S.~{Jo}, ``Deep full-body motion
  network for a soft wearable motion sensing suit,'' \emph{IEEE/ASME
  Transactions on Mechatronics}, vol.~24, no.~1, pp. 56--66, Feb 2019.

\bibitem{malinin2018predictive}
A.~Malinin and M.~Gales, ``Predictive uncertainty estimation via prior
  networks,'' in \emph{Advances in Neural Information Processing Systems},
  2018, pp. 7047--7058.

\bibitem{neal2012bayesian}
R.~M. Neal, \emph{Bayesian learning for neural networks}.\hskip 1em plus 0.5em
  minus 0.4em\relax Springer Science \& Business Media, 2012, vol. 118.

\bibitem{damianou2013deep}
A.~Damianou and N.~Lawrence, ``Deep gaussian processes,'' in \emph{Artificial
  Intelligence and Statistics}, 2013, pp. 207--215.

\bibitem{neuralprocess2018}
M.~Garnelo, J.~Schwarz, D.~Rosenbaum, F.~Viola, D.~J. Rezende, S.~Eslami, and
  Y.~W. Teh, ``Neural processes,'' \emph{ArXiv Preprint ArXiv:1807.01622},
  2018.

\bibitem{neal1996}
R.~M. Neal, ``Priors for infinite networks,'' in \emph{Bayesian Learning for
  Neural Networks}.\hskip 1em plus 0.5em minus 0.4em\relax Springer, 1996, pp.
  29--53.

\bibitem{matthews2018gaussian}
A.~G. d.~G. Matthews, M.~Rowland, J.~Hron, R.~E. Turner, and Z.~Ghahramani,
  ``Gaussian process behaviour in wide deep neural networks,'' \emph{ArXiv
  Preprint ArXiv:1804.11271}, 2018.

\bibitem{williams1997}
C.~K. Williams, ``Computing with infinite networks,'' in \emph{Advances in
  Neural Information Processing Systems}, 1997, pp. 295--301.

\bibitem{cho2009}
Y.~Cho and L.~K. Saul, ``Kernel methods for deep learning,'' in \emph{Advances
  in Neural Information Processing Systems}, 2009, pp. 342--350.

\bibitem{lee2017}
J.~Lee, Y.~Bahri, R.~Novak, S.~S. Schoenholz, J.~Pennington, and
  J.~Sohl-Dickstein, ``Deep neural networks as gaussian processes,''
  \emph{ArXiv Preprint ArXiv:1711.00165}, 2017.

\bibitem{pang2019}
G.~Pang, L.~Yang, and G.~E. Karniadakis, ``{Neural-net-induced Gaussian process
  regression for function approximation and PDE solution},'' \emph{Journal of
  Computational Physics}, vol. 384, pp. 270--288, 2019.

\bibitem{carroll2006}
R.~J. Carroll, D.~Ruppert, L.~A. Stefanski, and C.~M. Crainiceanu,
  \emph{Measurement error in nonlinear models: a modern perspective}.\hskip 1em
  plus 0.5em minus 0.4em\relax Chapman and Hall/CRC, 2006.

\bibitem{cervone2015}
D.~Cervone and N.~S. Pillai, ``Gaussian process regression with location
  errors,'' \emph{ArXiv Preprint ArXiv:1506.08256}, 2015.

\bibitem{zhu1998gaussian}
H.~Zhu, C.~K. Williams, R.~J. Rohwer, and M.~Morciniec, ``Gaussian regression
  and optimal finite dimensional linear models,'' in \emph{Neural Networks and
  Machine Learning}, C.~M. Bishop, Ed.\hskip 1em plus 0.5em minus 0.4em\relax
  Berlin: Springer-Verlag, 1998.

\bibitem{scikit-learn}
F.~Pedregosa, G.~Varoquaux, A.~Gramfort, V.~Michel, B.~Thirion, O.~Grisel,
  M.~Blondel, P.~Prettenhofer, R.~Weiss, V.~Dubourg, J.~Vanderplas, A.~Passos,
  D.~Cournapeau, M.~Brucher, M.~Perrot, and E.~Duchesnay, ``Scikit-learn:
  Machine learning in {P}ython,'' \emph{Journal of Machine Learning Research},
  vol.~12, pp. 2825--2830, 2011.

\bibitem{2020SciPy-NMeth}
P.~{Virtanen}, R.~{Gommers}, T.~E. {Oliphant}, M.~{Haberland}, T.~{Reddy},
  D.~{Cournapeau}, E.~{Burovski}, P.~{Peterson}, W.~{Weckesser}, J.~{Bright},
  S.~J. {van der Walt}, M.~{Brett}, J.~{Wilson}, K.~{Jarrod Millman},
  N.~{Mayorov}, A.~R.~J. {Nelson}, E.~{Jones}, R.~{Kern}, E.~{Larson},
  C.~{Carey}, {\.I}.~{Polat}, Y.~{Feng}, E.~W. {Moore}, J.~{Vand erPlas},
  D.~{Laxalde}, J.~{Perktold}, R.~{Cimrman}, I.~{Henriksen}, E.~A. {Quintero},
  C.~R. {Harris}, A.~M. {Archibald}, A.~H. {Ribeiro}, F.~{Pedregosa}, P.~{van
  Mulbregt}, and S.~.~. {Contributors}, ``{SciPy 1.0: Fundamental Algorithms
  for Scientific Computing in Python},'' \emph{Nature Methods}, vol.~17, pp.
  261--272, 2020.

\bibitem{wang2019effective}
Y.~Wang, X.~Yue, R.~Tuo, J.~H. Hunt, and J.~Shi, ``Effective model calibration
  via sensible variable identification and adjustment, with application to
  composite fuselage simulation,'' \emph{arXiv preprint arXiv:1912.12569},
  2019.

\bibitem{dace2018}
T.~Santner, B.~Williams, and W.~Notz, \emph{The Design and Analysis of Computer
  Experiments}.\hskip 1em plus 0.5em minus 0.4em\relax Springer, New York,
  2018.

\bibitem{yue2020active}
X.~Yue, Y.~Wen, J.~H. Hunt, and J.~Shi, ``Active learning for gaussian process
  considering uncertainties with application to shape control of composite
  fuselage,'' \emph{IEEE Transactions on Automation Science and Engineering},
  2020.

\end{thebibliography}


% \end{thebibliography}


% \begin{IEEEbiographynophoto}
% Biography Here.

%\newpage


%\vfill


\end{document}